\documentclass[10pt,conference]{IEEEtran}
\usepackage[utf8]{inputenc}

\usepackage[english]{babel}
\usepackage[letterpaper,top=2cm,bottom=2cm,left=3cm,right=3cm,marginparwidth=1.75cm]{geometry}
\usepackage{amsmath}
\usepackage{graphicx}
\usepackage{cite}
\usepackage[hyphens]{url}
\usepackage[colorlinks=true, allcolors=blue]{hyperref}
\usepackage{orcidlink}

\newcommand{\ignore}[1]{{}}

\title{LaaJMeter: A Framework for LaaJ Evaluation}

\author{
\IEEEauthorblockN{Samuel Ackerman\orcidlink{0000-0003-2631-0341}, Gal Amram\orcidlink{0000-0003-2138-7542}, Ora Nova Fandina, Eitan Farchi,\\ Shmulik Froimovich, Raviv Gal, Wesam Ibraheem, and Avi Ziv\orcidlink{0000-0002-6309-250X}}
\IEEEauthorblockA{IBM Research, Israel\\
Email: samuel.ackerman@ibm.com, gal.amram@ibm.com, ora.nova.fandina@ibm.com, farchi@il.ibm.com,\\ shmulik.froimovich@ibm.com, ravivg@il.ibm.com, wesam@il.ibm.com, aziv@il.ibm.com}

}

\begin{document}

	\maketitle
	%\sloppy
	
	\begin{abstract}
    
Large Language Models (LLMs) are increasingly used as evaluators in natural language processing tasks, a paradigm known as \emph{LLM-as-a-Judge} (LaaJ). The analysis of a LaaJ software, commonly refereed to as \emph{meta-evaluation}, pose significant challenges in domain-specific contexts. In such domains, in contrast to general domains, annotated data is scarce and expert evaluation is costly. As a result, meta-evaluation is often performed using metrics that have not been validated for the specific domain in which they are applied. Therefore, it becomes difficult to determine which metrics effectively identify LaaJ quality, and further, what threshold indicates sufficient evaluator performance.

In this work, we introduce \emph{LaaJMeter}, a simulation-based framework for controlled meta-evaluation of LaaJs. LaaJMeter enables engineers to generate synthetic data representing virtual models and judges, allowing systematic analysis of evaluation metrics under realistic conditions. This helps practitioners validate LaaJs for specific tasks: they can test whether their metrics correctly distinguish between high and low quality (virtual) LaaJs, and estimate appropriate thresholds for evaluator adequacy. 

We demonstrate the utility of LaaJMeter in a code translation task involving a legacy programming language, showing how different metrics vary in sensitivity to evaluator quality. Our results highlight the limitations of common metrics and the importance of principled metric selection. LaaJMeter provides a scalable and extensible solution for assessing LaaJs in low-resource settings, contributing to the broader effort to ensure trustworthy and reproducible evaluation in NLP.
\end{abstract}

\section{Introduction}

Large Language Models (LLMs)~\cite{NZY21, attention} have dramatically expanded the capabilities of Natural Language Processing (NLP), making a wide range of tasks such as translation, summarization, question answering, and code generation not only feasible but scalable and accessible~\cite{MK25, NKQSAUABM25}. Their ability to generalize across domains and adapt to diverse linguistic inputs has enabled the deployment of NLP tools in settings where traditional rule-based or analytical methods were previously impractical.

As NLP systems become increasingly pervasive, the need for reliable and efficient evaluation of their outputs grows more critical. Importantly, evaluating an NLP system, whether AI-based or not, is itself an NLP task. It involves interpreting textual outputs, assessing their correctness, relevance, and fluency, and making nuanced judgments that often depend on domain-specific expertise. This evaluative layer is essential for validating system behavior, guiding model development, and ensuring safe deployment~\cite{cambridge2023usereval}.

One common evaluation strategy employs text-similarity metrics such as BLEU~\cite{BLEU}, ROUGE~\cite{ROUGE}, METEOR~\cite{METEOR}, and BERTScore~\cite{BERTscore}, which compare system outputs against reference texts. While these metrics offer a convenient and automated means of assessing textual alignment, they often fail to capture the nuanced, context-dependent nature of evaluative tasks~\cite{BDMOS2022, CPL2020, EBSB23, saadany2021metrics}. BLEU and ROUGE rely heavily on n-gram overlap, which can penalize valid paraphrases or reward superficial similarity. BERTScore improves upon this by leveraging contextual embeddings, yet it still lacks the ability to evaluate reasoning quality, factual accuracy, or domain-specific relevance. These limitations are especially pronounced in tasks with multiple valid outputs or where subtle distinctions are critical. For instance, a prompt to generate code in a specific programming language may yield several correct solutions, none of which are textually similar.

A second approach uses off-the-shelf benchmarks that rely on analytic assessments or expert-annotated labels. HumanEval~\cite{HumanEval} evaluates code generation using unit tests. GLUE~\cite{GLUE} and SQuAD~\cite{SQuaD} employ human-annotated data to assess text understanding and reading comprehension, respectively. CNN/DailyMail~\cite{HermannKGEKSB15} and WMT~\cite{KocmiABBDFFFGGH24} use text similarity metrics like BLEU, ROUGE, and COMMET~\cite{ReiSFL20} against gold-standard human data to evaluate bullet summarization and translation, respectively (text similarity metrics are effective for these tasks, as correct outputs are expected to closely match the reference answers). Such benchmarks mainly exist for common NLP tasks.

When annotated data is unavailable, which is often the case for domain-specific tasks, a third approach involves recruiting experts to create dedicated datasets. While this method provides high-quality assessments, it is frequently prohibitively costly. Moreover, when specialized expertise is required, such as for translation or code generation in legacy languages, finding qualified annotators can be challenging. 

This paper focuses on a fourth approach. Given that evaluating NLP tools is itself an NLP task, and considering the remarkable success of LLMs in solving such tasks, there is growing interest in using LLMs for evaluation, a paradigm known as LLM-as-a-Judge (LaaJ)~\cite{CCL23}. This approach has proven effective and is increasingly adopted in both research and production environments~\cite{LIXWZ23, ZCSZWZLLLXZGS23, CWJSX23, HKPA24}. 
LaaJs are valued primarily for their scalability and cost-efficiency: they can provide context-aware evaluations across large datasets without requiring manual annotation or being limited to benchmark outputs. Furthermore, it avoids the risk of models being trained on the same benchmarks used for evaluation, a concern inherent in the second approach. Beyond standalone evaluation, LaaJs are also integrated into AI-agent architectures, where they assess the outputs of other components and influence downstream decisions. In such systems, the judgment of a LaaJ may directly affect the agent’s actions, making the reliability of these judgments critical.

However, the growing reliance on LaaJs introduces a critical question: who evaluates the evaluator? This challenge, known as \emph{meta-evaluation}, concerns the assessment of the judgments produced by an LLM acting as a judge \cite{chern2024scaleeval}. A robust meta-evaluation framework must determine whether these judgments are reliable, consistent, and meaningful across diverse tasks and inputs. A tempting but flawed solution is to use another LLM for this purpose. Yet doing so merely shifts the problem up a level, creating a recursive loop with no independent ground truth. Without an external, model-agnostic mechanism for evaluation, the validity of the entire process is called into question. Therefore, principled and reproducible approaches to meta-evaluation must break this dependency cycle and avoid relying on LLMs themselves.

For meta-evaluation, accessible benchmarks and human-annotated data are reused as follows~\cite{cambridge2023usereval, GJSTZXLSMLW24}. (1) The evaluated NLP tool is applied on benchmark or annotated data inputs; (2) the LaaJ assesses the tool outputs; (3) measures such as agreement, precision, recall, Cohen’s Kappa, and Spearman’s correlation are used to compare the alignment of LaaJ responses with human labels or benchmark results~\cite{BYCLHWYZXLZLH23, LZGSVKC24, TCRVH24}.

Consequently, shifting the evaluation task also shifts the underlying challenge: specialized tasks frequently suffer from data scarcity and limited access to expert annotations, due to the rarity and cost of qualified evaluators. This issue is especially pronounced in low-budget projects, where the resources needed to curate domain-specific datasets or engage experts are not readily available. As a result, in such contexts, meta-evaluation remains a significant hurdle, demanding creative solutions that compensate for the lack of standardized resources.

In the absence of standardized benchmarks and expert annotations, meta-evaluation in domain-specific contexts often relies on creative, task-specific strategies. One illustrative approach is to use outputs of known or controlled quality, such as responses from models with varying capabilities, and verify that the LaaJ assigns higher scores to the better outputs. However, due to the specificity of the domain, there is often little to no literature guiding which metrics are effective or which experimental setups yield meaningful insights for meta-evaluation \cite{hada2024metal}. Even when a metric appears to correlate with human judgment (though such correlation is rarely verifiable), determining a threshold that signifies sufficient evaluator quality remains inherently difficult. Without clear standards, practitioners are left to define ad hoc criteria, which may vary widely across applications and compromise reproducibility.

To address this challenge, we propose \emph{LaaJMeter}, a simulation-based framework for controlled meta-evaluation of LaaJs. LaaJMeter generates synthetic data representing virtual models and virtual LaaJs, enabling systematic analysis of evaluation metrics under realistic conditions. By simulating models of varying quality and LaaJs with different noise and bias profiles, we can test the sensitivity and robustness of metrics without relying on human annotation \cite{lu2025prompting}. In Section~\ref{sec:problem}, we illustrate the problem we address. In Section~\ref{sec:description}, we describe the LaaJMeter framework. 

We demonstrate the utility of LaaJMeter through a use case in code translation, where an LLM translates legacy code into a modern programming language.
The IBM watsonx Code Assistant for Z~\cite{WCA4Z} (WCA4Z) is IBM’s code assistant for software development on mainframe platforms. A distinctive feature of WCA4Z is its support for mainframe programming languages, such as the legacy language COBOL~\cite{COBOL}. A recently developed IBM service transforms COBOL code into the modern Java language~\cite{C2J}. 

As part of the evaluation of the COBOL translation to Java service, we developed several LaaJs and used the LaaJMeter framework to calibrate meta-evaluation of the LaaJs. Specifically, we simulate the metrics: \emph{t}-test, Kendall's $\tau$ correlation, and an ordering experiment. Our results reveal that while some metrics (\emph{t}-test) are insufficiently sensitive to LaaJ quality, others ($\tau$-correlation) provide robust signals even when model distance estimates are imprecise. The ordering experiment, though effective, requires careful interpretation of model distance to ensure meaningful results. We remark that we do not claim for the same metric efficiency in other domains. This is the core advantage of LaaJMeter: it allows to simulate metric simulation tailored to the specific domain, and thus to enhance confidence of the results. In section~\ref{sec:use-case} we describe the construction of the synthetic data for the use case. In Section~\ref{sec:simulation}, we present the LaaJMeter simulations for the use case and consequences. In Section~\ref{sec:distance}, we discuss distance estimations, and provide several suggestions for such estimations.

In summary, LaaJMeter offers a principled and extensible framework for evaluating LLMs as judges, particularly in settings where annotated data is scarce or unreliable. It complements existing evaluation paradigms and provides practical guidance for metric selection and thresholding. Our work contributes to the growing body of research on LLM-based evaluation and opens new directions for scalable, data-efficient assessment of model judgment quality.

\section{Related Work}
\label{sec:related-work}

Recent studies increasingly employ synthetic or semi-synthetic data to probe the reliability of LLM-as-a-Judge evaluators. Controlled simulation has emerged as a promising approach: researchers generate model responses with known or parametrically defined correctness, then assess whether a LaaJ reproduces the expected quality~\cite{kour2022measuringmeasuringtoolsautomatic,fandina2025automatedvalidationllmbasedevaluators}. This approach appears in several recent frameworks for evaluating factuality, reasoning, and preference alignment, where the ground truth is either human-defined or synthetically constructed~\cite{tan2025judgebenchbenchmarkevaluatingllmbased, fandina2025safesafetymetricautomatic}. Such works validate judges themselves, that is, they test whether a given LaaJ behaves as a reliable evaluator under known or controlled conditions.

When multiple judges are available, prior studies typically rely on conventional statistical decision metrics to select the best judge or to assess relative quality. Common approaches include mean-difference significance tests such as the t-test or ANOVA, correlation-based comparisons with baseline scores, inter-rater agreement statistics, and correlation with human judgments~\cite{sai2020surveyevaluationmetricsused,  chen2024humansllmsjudgestudy, dong2024llmpersonalizedjudge}. However, human-annotated data are often scarce or domain-specific, motivating the need for tailored techniques~\cite{fandina2025vintagecodemodernjudges} or even human-free methods to compare and select evaluators~\cite{chern2024l}.

 To our knowledge no prior work has validated the decision metrics themselves, namely, tested whether the statistical rules used to select among judges reliably identify the better evaluator under known ground-truth conditions.
Our framework provides the first controlled, label-free setting for such analysis. demonstrating, for example, that a simple t-test can yield significant results even for poor judges when model differences are large, and thus cannot serve as a reliable criterion for judge selection.

\section{Problem Description}
\label{sec:problem}

Domain-specific NLP tasks, particularly in low-budget projects, often suffer from a lack of high-quality human-labeled data. In such scenarios, evaluating the performance of a LaaJ becomes challenging due to the limited availability of reliable ground truth.

In the absence of a comprehensive expert evaluation, one way to assess LaaJ quality is by applying it to outputs of known, varying quality. The evaluated LaaJ is expected to correctly distinguish between different outputs to the same input, by assigning higher scores to better outputs. This form of meta-evaluation enables comparison between different LaaJs based on their ability to rank outputs in accordance with their true quality. See~\cite{fandina2025automatedvalidationllmbasedevaluators} for an example of such an evaluation. However, even when relative comparisons are possible, it remains difficult to define a satisfactory threshold for what constitutes acceptable LaaJ quality.

Another common scenario involves benchmarks annotated by a single expert. While such data can be used to compare LaaJs, for instance, by measuring their accuracy relative to the expert's scores, it remains unclear what level of agreement should be expected from a high-quality LaaJ. This stands in contrast to cases where multiple expert annotations are available, allowing for the calibration of LaaJ performance against inter-annotator agreement among human graders. In the absence of such comparative data, it becomes difficult to establish meaningful thresholds for acceptable LaaJ quality.

The core problem, therefore, is that when meta-evaluation data is scarce, it becomes difficult to determine:
\begin{itemize}
    \item Is a given metric suitable for evaluating LaaJ quality?
    \item What threshold should be considered acceptable for that metric?
\end{itemize}

\section{LaaJMeter, a General Description}
\label{sec:description}

To address the problem described above, we propose the creation of synthetic data representing virtual models and virtual LaaJs. This data is designed to simulate the behavior of models and LaaJs on a specific task, enabling controlled meta-evaluation. The key design principle is to ensure that the generated data closely resembles a realistic distribution and the behaviors observed in the evaluated NLP task. In Section~\ref{sec:use-case}, we present a use case that demonstrates the general approach outlined here. This construction example may help clarify the abstract concepts introduced in this section, which readers might otherwise find difficult to grasp.

\subsection{Virtual Points}  
We begin by defining a set of virtual points, which serve as inputs for the virtual models. These points are not instantiated with actual content; rather, we specify only the size of the set. A reasonable choice is to match the size of a typical benchmark used for the task under evaluation. The number of virtual points can influence the stability of metric calculations. For instance, larger sets tend to reduce the variance in the mean scores of evaluation metrics.

\subsection{Virtual Models}  
Once the number of virtual points is determined, we define virtual models. Conceptually, a virtual model produces inferences for each virtual point, which are then assigned ground-truth scores. In practice, we do not generate actual inferences, but rather assign scores directly. Formally, a virtual model is defined as a function $M: \{0, \ldots, n-1\} \rightarrow [a, b]$, where $n$ is the number of virtual points and $[a, b]$ is the range of possible scores. A good design choice is to ensure that the score distribution reflects typical patterns observed in the evaluated task.

\subsection{Additional Virtual Models}  
To simulate models of varying quality, we generate additional virtual models by systematically modifying the scores of an existing model. For example, given a model $M$, we can define a new model $M'$ such that $M'(k) = M(k) + a_k$, where the score adjustment $a_k$ is sampled from a distribution. If the distribution tends to be positive, $M'$ represents a model of higher quality than $M$. Importantly, this approach maintains a pointwise relationship between models, ensuring that $M'$ performs better than $M$ in expectation across all points (assuming $a_k$ has a positive expected value), mimicking the behavior of improved model versions.

\subsection{Virtual LaaJs}  
With virtual models defined, we proceed to construct virtual LaaJs. Intuitively, a virtual LaaJ evaluates the inferences produced by virtual models. Formally, a virtual LaaJ $L$ is a mapping $L(i, k) = s$, where $i$ is the index of a virtual model, $k$ is the index of a virtual point, and $s$ is the score assigned by the LaaJ to the model's inference on that point. The quality of a virtual LaaJ is determined by the expected deviation between its scores and the ground-truth scores of the virtual models. By generating LaaJs of varying quality, we can evaluate which metrics are effective in distinguishing between them.

\begin{table*}[h]
\centering
\begin{tabular}{c|cccccccccc}
\textbf{Distance/LaaJ} & $L_1$ & $L_2$ & $L_3$ & $L_4$ & $L_5$ & $L_6$ & $L_7$ & $L_8$ & $L_9$ & $L_{10}$ \\
\hline
1 & 0.07 & 0.08 & 0.11 & 0.19 & 0.20 & 0.20 & 0.21 & 0.25 & 0.24 & 0.29 \\
2 & 0.00 & 0.00 & 0.02 & 0.04 & 0.04 & 0.06 & 0.05 & 0.09 & 0.09 & 0.13 \\
3 & 0.00 & 0.00 & 0.00 & 0.01 & 0.01 & 0.01 & 0.02 & 0.01 & 0.02 & 0.05 \\
4 & 0.00 & 0.00 & 0.00 & 0.00 & 0.00 & 0.00 & 0.00 & 0.00 & 0.00 & 0.01 \\
5 & 0.00 & 0.00 & 0.00 & 0.00 & 0.00 & 0.00 & 0.00 & 0.00 & 0.00 & 0.00 \\
6 & 0.00 & 0.00 & 0.00 & 0.00 & 0.00 & 0.00 & 0.00 & 0.00 & 0.00 & 0.00 \\
7 & 0.00 & 0.00 & 0.00 & 0.00 & 0.00 & 0.00 & 0.00 & 0.00 & 0.00 & 0.00 \\
8 & 0.00 & 0.00 & 0.00 & 0.00 & 0.00 & 0.00 & 0.00 & 0.00 & 0.00 & 0.00 \\
9 & 0.00 & 0.00 & 0.00 & 0.00 & 0.00 & 0.00 & 0.00 & 0.00 & 0.00 & 0.00 \\
10 & 0.00 & 0.00 & 0.00 & 0.00 & 0.00 & 0.00 & 0.00 & 0.00 & 0.00 & 0.00 \\
\end{tabular}
\caption{$p$-value results}
\label{tbl:t-test}
\end{table*}

\section{Code Translation Use Case}
\label{sec:use-case}

The IBM watsonx Code Assistant for Z (WCA4Z) is IBM’s code assistant for software development on mainframe platforms. WCA4Z supports mainframe programming languages, such as the legacy language COBOL. A recently developed IBM service transforms COBOL code into the modern Java language (abbreviated as C2J). As part of the evaluation of the C2J service, we developed several LaaJs and used the LaaJMeter framework to assess their quality. In this section, we present our simulations and findings, thereby demonstrating an embodiment of the abstract concepts outlined in the previous section.

\subsection{Virtual Models}

We began by constructing a virtual base model $M_0: \{0,\dots, 99\} \rightarrow [0,30]$. This model was created by assigning ground-truth scores to point indices, sampled from the score distribution of a LaaJ we developed for the code translation evaluation. Accordingly, we chose the range $[0,30]$ to match the scoring scale used by that LaaJ. The number of points, 100, is typical to a benchmarks we use for model evaluation.

Following the construction of $M_0$, we generated additional virtual models $M_1, \dots, M_{20}$. Each model $M_{i+1}$ was derived from $M_i$ by modifying its scores as follows: for each point $k$, with probability $p$, we set $M_{i+1}(k) = \min\{M_i(k) + 1, 30\}$, incrementing the score unless it exceeds the maximum value of 30. With probability $1 - p$, we set $M_{i+1}(k) = \max\{M_i(k) - 1, 0\}$, decrementing the score unless it falls below 0. The value of $p$ was chosen such that the expected mean score of $M_{i+1}$ is 0.5 points higher than that of $M_i$, accounting for the number of scores at the boundaries (0 and 30).

We then constructed models with lower scores than $M_0$, denoted $M_{-1}, \dots, M_{-20}$, using the same procedure in reverse. For each $M_{i-1}$, the value of $p$ was selected so that its expected mean score is 0.5 points lower than that of $M_i$. In total, we created 41 virtual models: $M_{-20}, \dots, M_{20}$, with incrementally varying quality. For any pair $M_i$ and $M_j$ with $-20 \leq i < j \leq 20$, the mean score of $M_j$ exceeds that of $M_i$ by $\frac{j - i}{2}$. We refer to the value $j - i$ as the \emph{distance} between $M_i$ and $M_j$.

\subsection{Virtual LaaJs}
\label{subsec:virtual_laaj_use_case}

We now describe the construction of virtual LaaJs, informed by our experience evaluating LaaJs in the code modernization task. In practice, benchmarks often contain a mix of straightforward and challenging examples. A competent LaaJ typically performs well on simple cases, while its performance may vary on examples involving specific features, such as uncommon commands, that introduce evaluation difficulty. Some LaaJs may handle certain features effectively but struggle with others. Practical examples to such special features are the use of CICS commands, IMS commands, SQL queries etc.

To simulate this behavior, we designated 20 out of the 100 virtual points as ``simple points'', which all virtual LaaJs are expected to evaluate accurately, regardless of their quality. The remaining 80 points were divided into 10 disjoint sets of 8 points each, referred to as \emph{featured sets}. We then constructed virtual LaaJs $L_1, \dots, L_{10}$ as follows: for each LaaJ $L_j$, we randomly selected $j$ featured sets on which it performs poorly. On these sets, $L_j$ exhibits bias and high noise in its scoring; on all other points, it exhibits only low noise.

Formally, for a point $k$ not in the selected featured sets for $L_j$, the score is computed as $L_j(i, k) = M_i(k) + l_{i,k}$, where $l_{i,k} \sim \mathcal{N}(0, 1)$ represents low noise. Recall that $M_i(k)$ is the ground-truth score for model $M_i$ on point $k$.

If $k$ belongs to one of the featured sets selected for $L_j$, we sample a fixed bias $b_S \sim \mathcal{N}(0, 2)$ for the entire set $S$, and compute the score as $L_j(i, k) = M_i(k) + b_S + h_{i,k}$, where $h_{i,k} \sim \mathcal{N}(0, 5)$ represents high noise. This setup allows us to simulate LaaJs with varying sensitivity to specific features, enabling robust evaluation of metric effectiveness. Note that for $i < j$, $L_i$ outperforms $L_j$, as $L_j$ introduces both bias and high noise over an additional $j - i$ featured sets, corresponding to $8 \times (j - i)$ more points, compared to $L_i$.

\section{Code Translation Use Case: Metric Simulation and Findings}
\label{sec:simulation}

We present simulations for various metrics used in the meta-evaluation of LaaJs.

\subsection{$t$-Test}

\begin{table*}[t]
\centering
\begin{tabular}{c|cccccccccc}
\textbf{Distance/LaaJ} & $L_1$ & $L_2$ & $L_3$ & $L_4$ & $L_5$ & $L_6$ & $L_7$ & $L_8$ & $L_9$ & $L_{10}$ \\
\hline
1 & 0.79 & 0.76 & 0.71 & 0.67 & 0.65 & 0.60 & 0.57 & 0.56 & 0.53 & 0.48 \\
2 & 0.78 & 0.74 & 0.70 & 0.67 & 0.63 & 0.60 & 0.58 & 0.55 & 0.53 & 0.48 \\
3 & 0.76 & 0.73 & 0.69 & 0.65 & 0.63 & 0.59 & 0.56 & 0.56 & 0.52 & 0.47 \\
4 & 0.75 & 0.72 & 0.67 & 0.64 & 0.61 & 0.58 & 0.55 & 0.55 & 0.52 & 0.48 \\
5 & 0.74 & 0.71 & 0.67 & 0.63 & 0.62 & 0.57 & 0.55 & 0.55 & 0.52 & 0.46 \\
6 & 0.73 & 0.70 & 0.66 & 0.63 & 0.61 & 0.57 & 0.54 & 0.53 & 0.52 & 0.46 \\
7 & 0.72 & 0.69 & 0.65 & 0.62 & 0.61 & 0.56 & 0.53 & 0.53 & 0.50 & 0.46 \\
8 & 0.70 & 0.68 & 0.65 & 0.60 & 0.60 & 0.56 & 0.53 & 0.53 & 0.51 & 0.46 \\
9 & 0.69 & 0.67 & 0.63 & 0.60 & 0.59 & 0.55 & 0.53 & 0.53 & 0.50 & 0.44 \\
10 & 0.68 & 0.67 & 0.62 & 0.59 & 0.58 & 0.54 & 0.52 & 0.50 & 0.49 & 0.45 \\
\end{tabular}
\caption{Kendall-$\tau$ rank correlation coefficient results}
\label{tbl:tau}
\end{table*}

We begin with a negative result, one that demonstrates the limitations of a particular metric for meta-evaluation. Specifically, we assess LaaJ quality using a $t$-test applied to LaaJ scores over model inferences of varying quality.

The idea is as follows: we select two models (or model versions), where one is known to be better than the other. We apply a LaaJ to evaluate the inferences of both models and compute the $p$-value for the resulting score vectors. The expectation is that a high-quality LaaJ will yield a statistically significant result, indicating that the scores for the better model are higher. A poor LaaJ, however, may fail to detect this difference.

Importantly, the distance between the models also plays a critical role. If the distance is too large, even a poor LaaJ may detect the difference, leading to a significant $p$-value. To explore this, we simulated $t$-tests on virtual LaaJ scores across pairs of virtual models with varying distances. The resulting $p$-values are shown in Table~\ref{tbl:t-test}.

\subsubsection{$t$-Test Result Analysis and Conclusions}

As shown in Table~\ref{tbl:t-test}, all LaaJs yield statistically significant $p$-values for model distances greater than 2. This indicates that even a poor LaaJ, such as $L_{10}$, can detect differences between models, even when the distance is merely notable. Consequently, the $t$-test lacks sensitivity to LaaJ quality and fails to distinguish between LaaJs of varying performance. We conclude that the $t$-test is not suitable for meta-evaluation in this context.

\subsection{Kendall-$\tau$ rank correlation}

We now present a positive example, where our simulation supports the usefulness of a metric. The Kendall-$\tau$ rank correlation coefficient (also known as $\tau$-correlation) measures the agreement between two ranked vectors, with values ranging from $-1$ (complete disagreement) to $1$ (perfect agreement).

For meta-evaluation, we use $\tau$-correlation as follows: we select two models (or model versions) and apply them to the same benchmark. Then, we apply a LaaJ to evaluate the inferences and compute the $\tau$-correlation between the resulting score vectors. Since inference quality also depends on input complexity, we expect the ranking of scores to be preserved to some extent. A high-quality LaaJ should produce scores that maintain this ranking, resulting in a high $\tau$-correlation. In contrast, a poor LaaJ may fail to detect issues in lower-quality model outputs, leading to lower correlation.

Moreover, we expect the difference in $\tau$-correlation between good and poor LaaJs to be more pronounced when comparing models with small distances. In such cases, subtle differences in inference quality are harder to detect, and only better LaaJs are expected to capture them effectively.
The results of our simulation are presented in Table~\ref{tbl:tau}.

\subsubsection{$\tau$-Correlation Result Analysis and Conclusions}

The results align with our expectations. Better LaaJs consistently yield higher $\tau$-correlation scores, especially at smaller model distances. Notably, the metric is more sensitive to LaaJ quality than to model distance, making it particularly useful for meta-evaluation. This sensitivity allows practitioners to use $\tau$-correlation effectively without requiring precise knowledge of model distance.

Our findings suggest that a $\tau$-correlation score of approximately $0.70$ is sufficient to distinguish between high and low quality LaaJs, even when model distance is only roughly estimated. This insight supports the use of $\tau$-correlation as a reliable metric and helps establish a practical threshold for LaaJ evaluation. We conclude that $\tau$-correlation is an effective and robust metric for meta-evaluation of a C2J LaaJ.

\subsection{Ordering Experiment}

\begin{table*}[t]
\centering
\begin{tabular}{c|cccccccccc}
\textbf{Distance/LaaJ} & $L_1$ & $L_2$ & $L_3$ & $L_4$ & $L_5$ & $L_6$ & $L_7$ & $L_8$ & $L_9$ & $L_{10}$ \\
\hline
1 & 64.8\% & 64.6\% & 62.5\% & 62.4\% & 61.6\% & 61.1\% & 59.6\% & 59.5\% & 58.8\% & 57.2\% \\
2 & 74.3\% & 71.5\% & 70.6\% & 68.4\% & 67.1\% & 67.3\% & 65.3\% & 64.5\% & 63.4\% & 61.1\% \\
3 & 78.2\% & 77.0\% & 77.2\% & 74.7\% & 71.8\% & 70.9\% & 69.9\% & 68.3\% & 67.5\% & 63.9\% \\
4 & 83.2\% & 80.4\% & 79.7\% & 77.5\% & 76.3\% & 74.1\% & 73.1\% & 72.0\% & 70.8\% & 67.5\% \\
5 & 86.2\% & 84.6\% & 82.5\% & 81.1\% & 78.9\% & 76.7\% & 75.8\% & 73.9\% & 72.8\% & 69.8\% \\
6 & 88.4\% & 86.9\% & 85.5\% & 83.5\% & 82.3\% & 79.4\% & 78.5\% & 76.9\% & 75.7\% & 71.8\% \\
7 & 90.3\% & 88.9\% & 86.9\% & 86.0\% & 83.9\% & 81.7\% & 79.9\% & 78.6\% & 77.1\% & 74.4\% \\
8 & 91.6\% & 90.2\% & 89.2\% & 86.8\% & 86.3\% & 83.6\% & 82.3\% & 81.5\% & 81.1\% & 76.0\% \\
9 & 92.9\% & 91.3\% & 90.6\% & 88.4\% & 87.1\% & 85.1\% & 84.5\% & 82.0\% & 82.2\% & 77.6\% \\
10 & 94.4\% & 92.7\% & 92.2\% & 89.8\% & 88.4\% & 86.9\% & 85.5\% & 84.2\% & 83.7\% & 79.9\% \\
\end{tabular}
\caption{Ordering experiment results}
\label{tbl:order}
\end{table*}

In our final example, we demonstrate sensitivity to distance and thereby learn to apply a given metric appropriately. A test we frequently use for meta-evaluation is the ability of a LaaJ to correctly identify the better model among two candidates. To this end, we select two models of known, comparable quality, apply them to the same benchmarks, and compute the percentage of points for which the LaaJ (weakly) prefers the better model.

It is important to note that the better model does not necessarily produce superior inferences for all points. However, we expect that a more accurate LaaJ will yield higher results in this test, as it is less noisy, makes fewer mistakes, and is thus more likely to be biased toward the better model. We conducted a simulation of the ordering experiment using virtual models at various distances. The results are presented in Table~\ref{tbl:order}.

\subsubsection{Ordering Experiment Result Analysis and Conclusions} 
Our results show that the ordering experiment is sensitive to LaaJ quality and is therefore a useful metric for LaaJ meta-evaluation. However, the metric is also sensitive to model distance, which complicates the interpretation of results. A careful estimation of model distance is necessary to set a meaningful threshold for evaluation.

For example, a result of approximately $80\%$ is achieved by $L_3$ at distance 4, and also by the noisier $L_{10}$ at distance 10. This illustrates that without accounting for model distance, the metric may fail to distinguish between LaaJs of sufficent quality.

In summary, the ordering experiment is an effective metric, but its application requires careful handling. Specifically, an understanding of the relative distance between models.

\section{Distance Estimation}
\label{sec:distance}

The results presented in Section~\ref{sec:simulation} imply that, for some metrics, estimating model distance is necessary. The ordering experiment for C2J LaaJs is one example of this phenomenon. To support such estimation, we added an automated distance estimation component to the ordering experiments we perform. By running the experiment across several candidate LaaJs and two models of known, differing quality, we compute three estimates of model distance. Furthermore, for a given LaaJ and distance estimate, the automation identifies the virtual LaaJ among $L_1 \dots L_{10}$ whose performance is closest to that of the evaluated LaaJ (see Section~\ref{subsec:virtual_laaj_use_case} for the explicit construction).

The first distance estimate is the \emph{self-reference distance}: for a C2J LaaJ, this is defined as the difference between the mean scores assigned by the LaaJ to the two models. The second estimate is the \emph{average distance}: the average of all self-reference distances across the LaaJs used in the experiment. The third is the \emph{best performer distance}: the self-reference distance of the LaaJ that achieved the highest score in the ordering experiment. That is, the one that assigned a higher score to the second model for the greatest number of points.

The three options for distance estimations are intuitive methods our team is currently exploring. We do not claim that these are optimal, nor have we developed techniques to determine which among them is preferable. Developing methods to validate distance estimations and implementing them is beyond the scope of this paper and is left for future research.

\section{Conclusion}
\label{sec:conclusion}

In this work, we addressed the challenge of evaluating Large Language Models as Judges (LaaJs) in domain-specific NLP tasks where high-quality annotated data is scarce. We introduced the LaaJMeter framework, which enables controlled meta-evaluation through the construction of synthetic data representing virtual models and LaaJs. This approach allows for a systematic analysis of evaluation metrics under realistic conditions.

We reported a use case of using LaaJMeter for meta-evaluation of C2J LaaJs. A C2J laaj evaluates code translations of COBOL to Java, a service provided by IBM WCA4Z. Thtough the use case we demonstrated how LaaJMeter can simulate nuanced model and LaaJ behaviors. We evaluated several metrics, $t$-test, Kendall-$\tau$ correlation, and an ordering experiment, highlighting their strengths and limitations. Our findings show that while the $t$-test lacks sensitivity to LaaJ quality, Kendall-$\tau$ correlation provides a robust signal even with imprecise model distance estimates. The ordering experiment, although effective, requires careful calibration based on model distance.

\subsection{Future Research}

LaaJMeter is a framework designed to support meta-evaluation in scenarios where annotated data is scarce. It achieves this by applying metrics to virtual models and LaaJs, enabling analysis of metrics' ability to distinguish between high and low quality LaaJs. Essentially, the framework is valuable because it allows metric simulations to be run on synthetic data. This approach, using metric simulation to assess metric quality, is not exclusive to meta-evaluation. An important direction for future research is to generalize the construction of the virtual components (models and LaaJs, in our case) so that the core concepts of LaaJMeter can be applied to other domains as well.

A key factor in successful simulations using the LaaJMeter framework is the adequate construction of virtual models and LaaJs. For the C2J use case, we relied on experience, understanding, and intuition gained during the project to construct models that reflect reality as faithfully as possible. We wonder whether statistical tests could be developed to assess how well the virtual components reflect reality, for example, by observing LaaJ scores across various real benchmarks. Furthermore, it remains an open question whether virtual components could be constructed automatically, given LaaJ executions over benchmark datasets.

We noted that metrics may be sensitive to the distance between models, which we defined as the mean (ground truth) score difference for model outputs on the same input. Possible future research directions include validating whether a distance estimate is approximately correct and developing methods to automatically extract valid distance estimates. Since ground truth scores for models are not available, this task is inherently challenging. Moreover, precise distance estimation may be infeasible, and only approximate estimates can be expected.

% \bibliographystyle{plain}
% \bibliography{references}

\bibliography{References}
\bibliographystyle{IEEEtran}

\end{document}